%% file: cvpr.tex
\documentclass[final]{cvpr}

\usepackage{times}
\usepackage{epsfig}
\usepackage{graphicx}
\usepackage{amsmath}
\usepackage{amssymb}
\usepackage{stfloats}
\usepackage[title]{appendix}
\usepackage{verbatim}
\usepackage[pagebackref=true,breaklinks=true,colorlinks,bookmarks=false]{hyperref}
\usepackage{cleveref}
\usepackage{flushend}

\crefname{figure}{Fig.}{Fig.}
\Crefname{figure}{Figure}{Figures}
\crefname{equation}{}{}
\Crefname{equation}{Equation}{Equations}

\def\cvprPaperID{8453} 
\def\confYear{CVPR 2021}

\begin{document}
\title{Adaptive Consistency Regularization for Semi-Supervised Transfer Learning}

\author{Abulikemu Abuduweili$^{1,2}$\footnotemark[1], Xingjian Li$^{1,3*}$, Humphrey Shi$^{2\dag}$, Cheng-Zhong Xu$^{3}$, Dejing Dou$^{1\dag}$\\
\\
{\small $^1$Big Data Lab, Baidu Research, $^2$SHI Lab, University of Oregon,} \\{\small $^3$State Key Lab of IOTSC, Department of Computer Science, University of Macau}\\
{\small \{v\_abuduweili,lixingjian,doudejing\}@baidu.com, hshi3@uoregon.edu, czxu@um.edu.mo}
}




\maketitle
\thispagestyle{empty} 

\renewcommand{\thefootnote}{\fnsymbol{footnote}} \footnotetext[1]{Equal contributions and by alphabetical order. \dag ~Correspondence.}

\begin{abstract}
While recent studies on semi-supervised learning have shown remarkable progress in leveraging both labeled and unlabeled data, most of them presume a basic setting of the model is randomly initialized. In this work, we consider semi-supervised learning and transfer learning jointly, leading to a more practical and competitive paradigm that can utilize both powerful pre-trained models from source domain as well as labeled/unlabeled data in the target domain. To better exploit the value of both pre-trained weights and unlabeled target examples, we introduce \textbf{adaptive consistency regularization} that consists of two complementary components: Adaptive Knowledge Consistency (AKC) on 
the examples between the source and target model, and Adaptive Representation Consistency (ARC) on the target model between labeled and unlabeled examples. Examples involved in the consistency regularization are adaptively selected according to their potential contributions to the target task. We conduct extensive experiments on popular benchmarks including CIFAR-10, CUB-200, and MURA, by fine-tuning the ImageNet pre-trained ResNet-50 model. Results show that our proposed adaptive consistency regularization outperforms state-of-the-art semi-supervised learning techniques such as Pseudo Label, Mean Teacher, and FixMatch.  Moreover, our algorithm is orthogonal to existing methods and thus able to gain additional improvements on top of MixMatch and FixMatch. Our code is available at \href{https://github.com/SHI-Labs/Semi-Supervised-Transfer-Learning}{https://github.com/SHI-Labs/Semi-Supervised-Transfer-Learning}.
\end{abstract}

\input{sections/1_introduction}

\input{sections/2_relatedwork}
\input{sections/3_method}
\input{sections/4_experiment}

\section{Conclusion}
In this paper, we propose two regularization methods: Adaptive Knowledge Consistency (AKC) between the source and target model and Adaptive Representation Consistency (ARC) between labeled and unlabeled examples. We show that AKC and ARC are competitive among state-of-the-art SSL methods. Furthermore, by incorporating AKC and ARC with other SSL methods, we achieve the best performance among several baseline methods on various transfer learning benchmarks. Additionally, our adaptive consistency regularization methods could be used for more general transfer learning and (semi-) supervised learning frameworks. 

\section{Acknowledgements}

Parts of experiments in this paper were carried out on Baidu Data Federation Platform (Baidu FedCube). For usages, please contact us via \{fedcube,shubang\}@baidu.com.

{\small
\bibliographystyle{ieee_fullname}
\bibliography{egbib}
}

\input{sections/7_appendix}

\end{document}

%% file: sections/1_introduction.tex
\section{Introduction}

Deep neural networks have achieved great success in supervised learning tasks especially in computer vision~\cite{krizhevsky2012imagenet,he2016deep}. Yet, this heavily relies on a large amount of labeled data. As data annotation is usually expensive and time-consuming, Semi-Supervised Learning (SSL), which pursues the goal of effectively leveraging both labeled and unlabeled data, is widely studied. Recent state-of-the-art methods can be roughly summarized in three categories, which are consistency based regularization~\cite{laine2016temporal,tarvainen2017mean}, entropy minimization~\cite{grandvalet2005semi} and pseudo label~\cite{lee2013pseudo}.  

While most works focus on the general setting that training a randomly initialized model from scratch, we consider a more realistic setting utilizing the powerful pre-trained model which is adequately fit on large-scale datasets for general purposes such as ImageNet~\cite{deng2009imagenet} and Places365~\cite{zhou2017places}.  These pre-trained models are empirically proven to have excellent transferability on various down-streaming tasks~\cite{yosinski2014transferable} and can significantly improve the generalization capacity of target tasks especially when the sample size is relatively small. Moreover, they are free to fetch and can be efficiently fine-tuned to adapt to new tasks. A recent study~\cite{zhou2018semi} points out that the benefit of semi-supervised learning sometimes may be marginal when fine-tuning a pre-trained model on the target dataset.  However, the investigation of a systematic solution on DNN-based semi-supervised transfer learning has rarely been delved into.

In this work, we propose a semi-supervised transfer learning framework beyond a simple combination of these two kinds of algorithms. We extend the effective idea of consistency regularization in semi-supervised learning to adapt to inductive transfer learning, where the pre-trained weight learned by the source task is available. Specifically, our method is composed of two essential components: (1) Adaptive Knowledge Consistency (AKC) on 
the examples between the source and target model. We utilize  target examples to transfer knowledge from the pre-trained model and help generalize the target model inspired by recent studies about knowledge distillation~\cite{zagoruyko2016paying} and transfer learning~\cite{li2018explicit}. To cope with the risk of negative transfer~\cite{torrey2010transfer} caused by the discrepancy between the source and target task, we use the knowledge adaptive sample importance for proper cross-task knowledge consistency regularization. Intuitively, we are inclined to select examples lying in the trusted region of the source model.  (2) Adaptive Representation Consistency (ARC) on the target model between labeled and unlabeled examples. In transfer learning applications, labeled examples are often insufficient and thus they are prone to be projected onto an inappropriate representation with only the supervision of their labels. To tackle this problem we utilize ample unlabeled examples to adjust the representation produced by supervised learning to the real target domain. This is achieved by minimizing their Maximum Mean Discrepancy (MMD) distance. Furthermore, we adaptively decide the sample set used for restricting the representation distance. An intuitive explanation about the motivation of ARC is showed in supplementary A.

We evaluate our method on several semi-supervised transfer learning settings considering various typical scenarios. We use popular datasets CIFAR-10, CUB-200-2011, MIT Indoor 67, and MURA,
covering domains including objects, animals, scenes and, radiographs. 

Our main contributions can be summarized in the following points. 
\begin{itemize}
\vspace{-2mm}
\setlength{\itemsep}{0pt}

\item To the best of our knowledge, we are the first to propose an advanced end-to-end semi-supervised transfer learning framework for deep neural networks. Considering incorporating inductive transfer learning, our research is closer to the actual problems in practice. Previous empirical study~\cite{zhou2018semi} provided observations and understandings by directly combining SSL with fine-tuning, but did not develop effective algorithms. 
\item  We introduce adaptive consistency regularization to improve semi-supervised transfer learning by exploiting the characteristics of both semi-supervised learning and transfer learning, including cross-task knowledge distillation with adaptive sample importance named Adaptive Knowledge Consistency and representation adaptation for supervised learning using selected unlabeled data as the reference named Adaptive Representation Consistency. 
\item  We conduct extensive experiments and show that the proposed adaptive consistency regularization is superior to classic semi-supervised learning algorithms such as Pseudo Label, Mean Teacher, and MixMatch on various semi-supervised transfer learning tasks. Furthermore, our method is shown orthogonal to existing methods and can obtain additional improvements even on top of MixMatch and FixMatch, which combine several state-of-the-art SSL techniques. 
\end{itemize}

%% file: sections/2_relatedwork.tex
\section{Related Work}
\subsection{Deep Transfer Learning}
 Previous research \cite{pan2009survey} proposed a comprehensive survey dividing transfer learning into three categories, which are inductive transfer learning, transductive transfer learning, and unsupervised transfer learning, according to the relationship between the source and target domain, and whether examples are labeled in either domain. In the deep learning community, most concerned transfer learning tasks include fine-tuning, domain adaptation, and few-shot learning. In this paper, we focus on fine-tuning as the main method, which belongs to inductive transfer learning according to ~\cite{pan2009survey}. 

\textbf{Fine-tuning}. Previous research pointed out that deep neural networks well-trained on large scale datasets for general purpose show great transferability on various downstream tasks~\cite{yosinski2014transferable}. Thus fine-tuning a pre-trained model to adapt new tasks has become a popular paradigm for many real world applications~\cite{huh2016makes}. To further improve the effectiveness, some methods are investigated to improve the knowledge exploitation of the pre-trained model during fine-tuning, instead of merely treating it as a better starting point than random initialization. For example, ~\cite{li2018explicit} argued that the starting point should be used as the reference to regularize the learned weight. ~\cite{zagoruyko2016paying} demonstrated that knowledge distillation through attention map can be applied to different tasks and useful to enhance the performance of transfer learning. ~\cite{li2019delta} proposed a channel level attention for knowledge distillation from the source to target task. Besides the idea of utilizing the pre-trained model, there are studies from other perspective, such as sample selection~\cite{ge2017cvpr,cui2018large,ngiam2018domain}, dynamic fine-tuning path selection~\cite{zhang2018parameter,guo2019spottune} and suppressing negative transfer~\cite{wan2019towards,li2020rifle}.

\textbf{Domain Adaption}. Different from fine-tuning, domain adaptation~\cite{saenko2010adapting} copes with the problem of sample selection bias between the training and test data. An important concept in classic domain adaptation methods is to generate domain invariant representation over the training set. Some earlier studies~\cite{gong2013connecting,huang2007correcting} proposed sample re-weighting algorithms to adjust the decision boundary learned by the training examples to adapt to the target domain. Another useful idea is to explicitly minimize the distribution distance between the source and target domain. This kind of methods~\cite{pan2010domain,long2015learning,wang2020alleviating} intend to learn a proper feature transformation that can simultaneously project both domains into a shared representation space. Our work is highly inspired by the critical ideas developed for domain adaptation such as sample re-weighting and representation adaptation, while the task is rather different. 

\textbf{Few-shot Learning}. Few-shot learning has been paid to increasing attention in recent years as it aims at imitating human intelligence by which knowledge can be generalized provided only several examples. The mainstream research direction is related to meta learning~\cite{finn2017model,snell2017prototypical}. It is quite different from regular transfer learning paradigms that the transferred knowledge is how to learn rather than what (e.g. model parameter) has learned. Recent work~\cite{yu2020transmatch} designed a semi-supervised few-shot learning framework TransMatch by incorporating Imprinting and MixMatch. They demonstrated that utilizing unlabeled examples makes their framework surpass the purely supervised few-shot learning competitors. 

\subsection{Semi-Supervised Learning}
There exist a vast number of classic works on semi-supervised learning, and most of them fall into one of the three main mechanisms\cite{oliver2018realistic}: consistency based regularization, entropy minimization, and pseudo label. All these methods share an intuition to use additional unlabeled data to exploit the underlying structure, which usually could hint the separation of samples whose labels we want to distinguish. We only briefly discuss the branch of consistency based regularization, which is the most related to our work. 

Consistency regularization is based on the hypothesis that the decision boundary is not likely to pass through high-density areas. This hypothesis results in a specific principle that a sample and its close neighbours are expected to have the same label. This forms the basic motivation of consistency  based methods, as well as many self-supervised learning approaches, which all care about the utilization of unlabeled data. For example, the $\Pi$ model~\cite{laine2016temporal} arguments the input sample with different noises, and adds a regularization term to reduce the discrepancy between outputs with respect to the original input and its perturbed  peers. Temporal Ensembling~\cite{laine2016temporal} and Mean Teacher~\cite{tarvainen2017mean} involve ensemble learning to promote the quality of labels of the perturbed samples. Specifically, they use the moving average weights or predictions. Recently, Interpolation Consistency Training (ICT)~\cite{verma2019interpolation} improved the perturbation method by using Mixup with another unlabeled sample instead of adding  random noise. This is regarded as a more efficient transformation when dealing with low-margin unlabeled points. MixMatch~\cite{berthelot2019mixmatch} further proposed artificial label sharpening for unlabeled data and mixing both labeled and unlabeled data in Mixup. FixMatch~\cite{sohn2020fixmatch} continued the trend to combine diverse mechanisms for exploiting unlabeled examples.

Our work does not pursue to search for the best choice among those general semi-supervised learning algorithms in the transfer learning setting. Instead, we intend to develop more targeted strategies utilizing the properties of the combination of semi-supervised and transfer learning problems. 

\subsection{Semi-Supervised Transfer Learning}
Semi-supervised transfer learning can be regarded as a natural extension of regular semi-supervised learning by taking a related auxiliary task into consideration or as an extension of regular transfer learning with only a proportion of the labeled target examples. There are few works targeting this sort of problem. Early work~\cite{shi2009extending} investigated this problem under the setting of the traditional machine learning framework. They proposed an improved co-training method for inductive transfer learning with instance re-weighting according to the training error. Two diverse k-Nearest-Neighbour (kNN) learners with different values of k are trained collaboratively. Recently, ~\cite{zhou2018semi} presented an empirical study showing that the gains from state-of-the-art SSL techniques decrease or sometimes even disappear compared with a fully-supervised baseline when we fine-tune the target task starting from a pre-trained model. While these observations pointed out the necessity of considering this more competitive and practice baseline, they did not aim at inventing a solution. ~\cite{jakubovitz2019lautum} imposed the Lautum regularization with which they improved the pre-training stage using examples from both the source and target task. Although the accuracy outperforms several baselines, the requirements of accessing the source dataset and an extra pre-training for every target task are usually unrealistic. 

Some recent studies investigated semi-supervised transfer learning on specific tasks. ~\cite{wei2019semi} discussed the task of rain removal with a framework of semi-supervised transfer learning. \cite{wang2020alleviating} introduced a semi-supervised domain adaptation method for semantic segmentation. \cite{fu2019self} studied pseudo-labeling method on unsupervised domain adaptation for person re-identification.

Different from those works, this paper introduces a novel framework for general semi-supervised transfer learning. 

%% file: sections/3_method.tex
\begin{figure*}[t]
\begin{center}
\includegraphics[height=0.43\textwidth,width=0.9\textwidth]{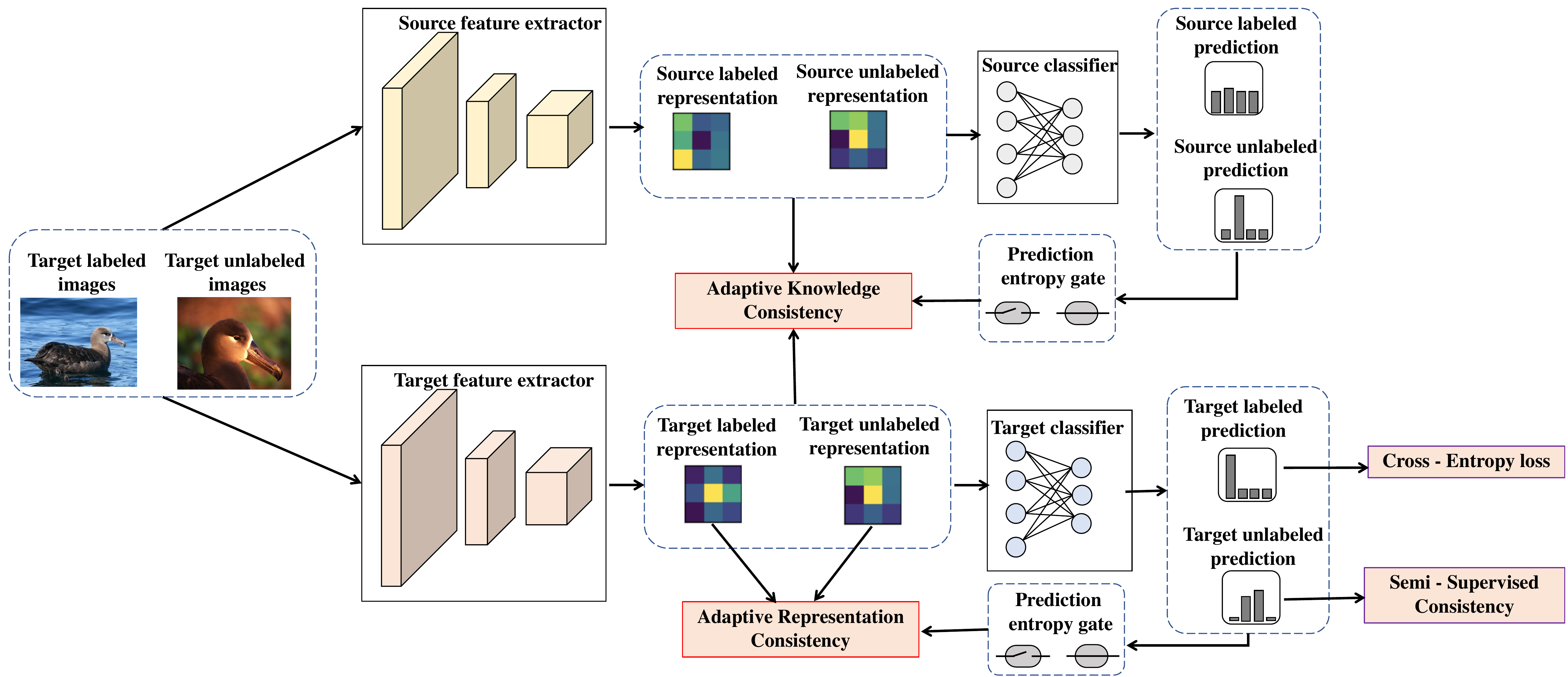}
\caption{The framework of adaptive consistency regularization for semi-supervised and transfer learning.}
\label{fig:arch}
\end{center}
\vspace{-5mm}
\end{figure*}

\section{The Proposed Framework}
The flowchart of the proposed semi-supervised transfer learning is illustrated in \Cref{fig:arch}. Please check the more detailed illustration of the proposed framework in supplementary A.

\textbf{Problem definition:} In inductive transfer learning, we have the source dataset $\mathcal{D}_s$ and the target dataset $\mathcal{D}_t$ corresponding to different tasks. A typical deep neural network $f$ can be split into two parts: a representation function $F_{\mathbf{\theta}}$ and a task-specific function $G_{\mathbf{\phi}}$. $F_{\mathbf{\theta}}$ is able to contain general knowledge if trained over a dataset with diverse semantics and thus is transferable. While $G_{\mathbf{\phi}}$ has the particular architecture with respect to the task attribute such as the number of classes. We denote the parameters of the representation function (called feature extractor in our task) and task-specific function (called classifier in our task) pre-trained over the source dataset as $\mathbf{\theta}^0$ and $\mathbf{\phi}^0$ respectively. For the target dataset, we denote $\mathcal{D}_t^l = \{\mathbf{x}_l^{1,2…n}\}$ as the labeled examples and $\mathcal{D}_t^u = \{\mathbf{x}_u^{1,2…,m}\}$ as the unlabeled examples. Here we ignore the subscript $s$ or $t$ for a specific example $\mathbf{x}$ as we will only use the target dataset after the pre-training stage. 
We then define the complete target dataset $\mathcal{D}_t = \mathcal{D}_t^l \cup \mathcal{D}_t^u $ and its size is $n+m$. To solve the target task, we formalize the general form of the optimizing objective as
\begin{equation}
    \mathbf{\theta}^*, \mathbf{\phi}^* = \mathop{\arg\min}_{\mathbf{\theta},\mathbf{\phi}} \sum_{i=1}^n L_{\rm CE}(\mathbf{\theta},\mathbf{\phi};\mathbf{x}_l^i) + R(\theta)
\end{equation}
, where $L_{\rm CE}$ is the commonly used cross-entropy loss indicating the prediction error and $R$ refers to additional regularization related to the pre-trained parameter $\theta^0,\phi^0$ and the target dataset $\mathcal{D}_t$. Note that since a labeled example can be regarded as unlabeled if we ignore its label, we actually use $\mathcal{D}_t$ when we need a set of unlabeled examples.

\subsection{Pre-training and Imprinting}
We adopt a popular strategy to implement inductive transfer learning, which is to sequentially learn from the source and the target dataset. The first step is pre-training. The representation parameter of the target model is initialized with $\mathbf{\theta}^0$. We do not discuss other paradigms of utilizing the source dataset in this paper such as co-training the source and target dataset like~\cite{ge2017cvpr}.

Although the task-specific function $G$ can not be shared directly, we borrow the idea of Imprinting from recent low-shot learning research~\cite{qi2018low}. Imprinting performs an informative initialization on $G$ instead of random initialization. Such knowledge derived from the feature extractor $F$ of the source model provides a much better starting point to the target model with immediate good classification performance.

\subsection{Adaptive Knowledge Consistency}
Knowledge distillation is widely studied with the original motivation of compressing complex ensembling models~\cite{HintonDistilling}. While recent studies reveal that knowledge distillation can also help improve the identical model~\cite{furlanello2018born} over the same task and even generalize a different task~\cite{zagoruyko2016paying,li2017learning,li2019delta}. We adopt the method to distill the knowledge of the source model through the representation rather than the task-specific logits output, as the latter is not suitable for handling different tasks. While different from previous studies, we employ both labeled and unlabeled data as the bridge of knowledge transfer and impose adaptive sample importance to prevent negative transfer cause by the discrepancy between the two datasets. Specifically, we constrain the weighted  Kullback–Leibler divergence (or mean square error) of outputs between the pre-trained feature extractor $F_{\mathbf{\theta^0}}$ and the target feature extractor $F_{\mathbf{\theta}}$ using the entire target dataset $\mathcal{D}_t$. In our setting, we denote $\mathcal{L}= \{\mathbf{x}^i_l \}^{B_l} \subset \mathcal{D}_t^l$ as a mini-batch of ${B_l}$ labeled examples, and $\mathcal{U}= \{\mathbf{x}^i_u\}^{B_u}  \subset \mathcal{D}_t^u$ as a mini-batch of ${B_u}$ unlabeled examples. Formally, the regularization term of a mini-batch can be written as 
\begin{equation}
    R_{\rm K} = \frac{1}{B_l + B_u} \sum_{\mathbf{x}^i \in \mathcal{L} \cup \mathcal{U} } w^i_{\rm K} \ \mathrm{KL}(F_{\mathbf{\theta}^0}(\mathbf{x}^i), F_{\mathbf{\theta}}(\mathbf{x}^i))
\end{equation}

To calculate the sample importance $w^i_{\rm K}$, we leverage the pre-trained source model with the parameter $\mathbf{\theta}^0$ and $\mathbf{\phi}^0$. In detail, a target example $\mathbf{x}^i$ is fed forward the pre-trained model and we obtain the final output post-processed by the softmax operation, marked as $\mathbf{p}^i_s = G_{\phi^0}(F_{\theta^0}(\mathbf{x}^i))$. $\mathbf{p}^i_s$ is a 1-dimensional vector with the length equal to the number of source classes $C_s$. We get the weight of sample $\mathbf{x}^i$ by calculating the entropy of $\mathbf{p}^i_s$ as:
\begin{equation}
w^i_{\rm K} = \mathcal{G}( \mathrm{H}(\mathbf{p}^i_s)) = \mathcal{G} (-\sum_{j=1}^{C_s} \mathbf{p}^i_{s,j} {\rm log}(\mathbf{p}^i_{s,j})).
\end{equation}
Where $\mathcal{G}$ is an entropy-gate function, which projects calculated entropy to a value of sample importance.  Intuitively, the entropy of the output as a probability on different classes indicates the confidence of the recognition with respect to the input. In other words, higher output confidence implies that the input sample is more likely to fall into the source model's trust region and consequently the knowledge about this sample is reliable to the target model. In our implementation, we perform a hard filter according to the sample importance with a pre-determined threshold value $ \epsilon_{\rm K}$ so as to reduce the extra computation burden. Sample importance $w^i_{\rm K}$ can be written as a binary value of:
\begin{equation}
w^i_{\rm K} = \mathrm{I}( \mathrm{H}(\mathbf{p}^i_s) \le \epsilon_{\rm K})
\end{equation}
The sample importance $w^i_{\rm K} = 1$, only if the corresponding entropy is lower than pre-determined threshold $\mathrm{H}(\mathbf{p}^i_s) \le \epsilon_{\rm K}$.

\subsection{Adaptive Representation Consistency}
In this part, we introduce another imposed regularizer named adaptive representation (distribution) consistency, by which we intend to tackle the problem of over-fitting the insufficient labeled target samples. Motivated by the fact that unlabeled samples themselves contain potential information about the data structure, we utilize unlabeled target samples to help labeled samples learn representations with stronger generalization ability. Different from knowledge distillation incorporating the alignment at the sample level, the representation consistency affects training at the distribution level. Specifically, we use the classical metric Maximum Mean Discrepancies (MMD)~\cite{borgwardt2006integrating} to measure the distance between the representations of labeled and unlabeled data. 
Denoting $\mathbf{V}=\{\mathbf{v}^1, \mathbf{v}^2,...,\mathbf{v}^n\}$ and $\mathbf{U}=\{\mathbf{u}^1, \mathbf{u}^2,...,\mathbf{u}^m\}$ as random variable sets with distributions $Q_v$ and $Q_u$, an unbiased estimate of the MMD between $Q_v$ and $Q_u$ compares the square distance between the empirical kernel mean embeddings as
\begin{equation} 
\begin{aligned}
\mathbf{MMD}(Q_v, Q_u) & = \| \frac{1}{m}\sum_{i=1}^m \kappa(\mathbf{v}^i) - \frac{1}{n}\sum_{j=1}^n \kappa(\mathbf{u}^j) \|^2,
  \end{aligned}
\end{equation}
where $\kappa$ refers to the kernel, as which a Gaussian radial basis function (RBF) is usually used in practice~\cite{gretton2012kernel,long2017deep}. 


In our case, we need to measure the MMD between labeled representation $\{F_\theta(\mathbf{x}^i_l) | \mathbf{x}^i_l \in \mathcal{L} \}$ distribution and unlabeled representation $\{F_\theta(\mathbf{x}^i_u) | \mathbf{x}^i_u \in \mathcal{U} \}$ distribution. Nevertheless, this restrain raises a severe risk because the target model is progressively learned. Thus even the representation distribution obtained by sufficient unlabeled examples is inaccurate at earlier stages of the training procedure. To overcome this kind of problem, we involve an adaptive sample selection method similar to that in adaptive knowledge consistency. Specifically, we compute the entropy of the softmax output given a sample as the input and regard the entropy as the target model's confidence on this sample. Only confident samples will be employed to regularize the representation of labeled data. In detail, a labeled example $\mathbf{x}^i_l$ (and an unlabeled example $\mathbf{x}^i_u$) is fed forward the target model and we obtain the final output as $\mathbf{p}^i_l = G_{\phi}(F_{\theta}(\mathbf{x}^i_l))$ (and $\mathbf{p}^i_u = G_{\phi}(F_{\theta}(\mathbf{x}^i_u))$), then we get the gate state (whether selection or not) of the example  by calculating the entropy of prediction as $\mathbf{H}(\mathbf{p}^i_l)$ (and $\mathbf{H}(\mathbf{p}^i_u)$) considering predefined threshold value $\epsilon_{\rm R}$. Denoting set of selected labeled representation as $\mathcal{F}_l$ and set of selected unlabeled representation as $\mathcal{F}_u$:

\begin{equation} 
\begin{aligned}
& \mathcal{F}_l = \{ F_\theta(\mathbf{x}^i_l) | x^i_l \in \mathcal{L} ~{\rm and}~ \mathbf{H}(\mathbf{p}^i_l) \le \epsilon_{\rm R} \} \\
& \mathcal{F}_u = \{ F_\theta(\mathbf{x}^i_u) | x^i_u \in \mathcal{U} ~{\rm and}~ \mathbf{H}(\mathbf{p}^i_u) \le \epsilon_{\rm R} \} 
  \end{aligned}
\end{equation}

Note that the sample selection result is adaptively changing as the target model progressively fits more training examples. Considering that the number of selected samples in a mini-batch may not be adequate to calculate a convinced distribution, we impose a replay buffer to save recent selected confident examples. The replay buffer enables us to calculate MMD with more data, and which is helpful to approximate full representation distribution with recent some mini-batches representation distribution. 
The pseudo-code of the replay buffer is quite straightforward, as following:
\begin{equation} 
\begin{aligned}
& {\rm Labeled\_Buffer.update}(\mathcal{F}_l)\\
& {\rm Unlabeled\_Buffer.update}(\mathcal{F}_u) \\
& \mathcal{F}_l^{\star} = {\rm Labeled\_Buffer.get\_last\_k}() \\
& \mathcal{F}_u^{\star} = {\rm Unlabeled\_Buffer.get\_last\_k}() 
\end{aligned}
\end{equation}

Denoting $Q_{\mathcal{F}_l^{\star}}$ and $Q_{\mathcal{F}_u^{\star}}$ as the representation distribution generated from $\mathcal{F}_l^{\star}$ and $\mathcal{F}_u^{\star}$, we give the adaptive representation consistency as the following form:
\begin{equation}
    R_\mathbf{R} = \mathbf{MMD}(Q_{\mathcal{F}_l^{\star}}, Q_{\mathcal{F}_u^{\star}}).
\end{equation}

\subsection{Summarization of the Framework}
We finally present the complete adaptive consistency regularization consisting of AKC and ARC as
\begin{equation}
    R(\mathbf{\theta}) = \lambda_\mathbf{K} R_\mathbf{K} + \lambda_\mathbf{R} R_\mathbf{R}.
\end{equation}
Where $\lambda_\mathbf{K}$ and $\lambda_\mathbf{R}$ are weighted factors for AKC and ARC.
If we incorporate cross-entropy loss $L_{\rm CE}$ for labeled data and semi-supervised consistency loss $L_\mathbf{S}$ for unlabeled data (just like MixMatch, FixMatch, Pseudo-labeling ...), then the final loss function would become:
\begin{equation}
\begin{aligned}
    L(\theta,\phi) =&& \frac{1}{n} \sum_{i=1}^n L_{\rm CE}(\mathbf{\theta},\mathbf{\phi};\mathbf{x}_l^i) +  
    \lambda_\mathbf{S} L_\mathbf{S}(\{\mathbf{x}_u^i\}) + \\
  &&  \lambda_\mathbf{K} R_\mathbf{K}(\{\mathbf{x}_l^i\},\{\mathbf{x}_u^i\}) + \lambda_\mathbf{R} R_\mathbf{R}(\{\mathbf{x}_l^i\},\{\mathbf{x}_u^i\})
\end{aligned} 
\end{equation}
Where $\lambda_\mathbf{S}$ is a weighted factor for semi-supervised consistency loss.
After initializing with the pre-trained source model and imprinting, the remaining fine-tuning is performed in an end-to-end manner.  

%% file: sections/4_experiment.tex
\section{Experiments}

\subsection{Experimental setup}

\subsubsection{Dataset configuration}

We evaluate our proposed adaptive consistency regularization methods and compare with state-of-the-art semi-supervised learning methods on several public datasets including the commonly used semi-supervised learning dataset CIFAR-10~\cite{krizhevsky2009learning} and transfer learning benchmarks CUB-200-2011\cite{wah2011caltech}, MIT Indoor-67\cite{quattoni2009recognizing} and musculoskeletal radiographs dataset MURA\cite{rajpurkar2017mura}. ImageNet\cite{deng2009imagenet} is used as the source task. Note that CIFAR-10, Indoor-67 and CUB-200-2011 have some classes semantically overlaps with ImageNet, while MURA is a medical image dataset with a large domain mismatch from ImageNet. Detailed descriptions about these datasets are listed in supplementary  B.1.  

\subsubsection{Baseline}
We compare proposed adaptive consistency regularization methods with the following state-of-the-art semi-supervised learning methods. In order to make a fair comparison in semi-supervised transfer learning tasks, we incorporate these semi-supervised learning methods with the same strategies including initialization with imprinting and fine-tuning all layers.

\begin{itemize}
\vspace{-2mm}
\setlength{\itemsep}{0pt}

\item Standard fine-tuning on labeled dataset: This is equivalent to a pure supervised manner where unlabeled examples are not used. 

\item Pseudo-labeling~\cite{lee2013pseudo}: It proceeds by producing “pseudo-labels” for unlabeled training set using the prediction function itself over the course of training. 

\item Mean-teacher~\cite{tarvainen2017mean}: It obtains more stable target predictions for unlabeled training set. Specifically, it  sets the target labels using an exponential moving average of parameters from previous training steps. The representation consistency between the original and perturbed unlabeled samples is encouraged, as well as the standard cross-entropy minimization for labeled samples. 

\item MixMatch~\cite{berthelot2019mixmatch}: In addition to the consistency regularization, it proposes artificial label sharpening for pseudo-labeling on unlabeled data and mixing both labeled and unlabeled data in Mixup during the process of fine-tuning.

\item FixMatch~\cite{sohn2020fixmatch}: FixMatch further improves on top of the above techniques. It computes an artificial label given a weakly augmented version of a given unlabeled image. Then it uses the pseudo-label to enforce the cross-entropy loss against the model’s output for a strongly-augmented version of the unlabeled image.
\end{itemize}

It should be noted that our proposed adaptive consistency regularization techniques are theoretically compatible with other semi-supervised methods. Thus, we also evaluate our proposed regularization techniques integrated with MixMatch or FixMatch.

\subsubsection{Training strategy}
On the transfer learning benchmarks, we use ImageNet as our source dataset and use ResNet-50\cite{he2016deep} pre-trained model as our source model by default unless explicitly specified. 
We fine-tune the ImageNet pre-trained model on CUB-200-2011, Indoor-67, and MURA datasets with labeled and unlabeled samples. We use SGD with momentum as the optimizer to train the target model 200 epochs. The momentum rate is set to be 0.9, the initial learning rate is 0.001 (except that the initial learning rate is 0.01 for CUB-200-2011) and the mini-batch size is 64 for both labeled and unlabeled dataset. For a learning rate schedule, we use a cosine learning rate decay\cite{loshchilov2016sgdr} which sets the learning rate to
\begin{equation}
\eta_t = \eta_0{\rm cos}(\frac{7 \pi t}{16 T})
\end{equation}
where $\eta_0$ is the initial learning rate, $t$ is the current training step, and $T$ is the total number of training steps. For our semi-supervised fine-tuning method, we set the parameters of AKC and ARC as follows. We set the regularization weight factors as $\lambda_\mathbf{K}=1$ and $\lambda_\mathbf{R}=30$, and adaptive thresholds as $\epsilon_\mathbf{K}=0.7 \cdot {\rm log}(C_s)$ and $\epsilon_\mathbf{R}=0.7 \cdot {\rm log}(C_t)$. Where $C_s$ and $C_t$ refer to the class number of source dataset and target dataset.

On the CIFAR-10 experiment, following the experiment setting by \cite{sohn2020fixmatch}, we use the same network architecture Wide ResNet-28-2 \cite{zagoruyko2016wide} and training protocol, including the optimizer, learning rate schedule, data preprocessing, across all SSL methods. In the pre-training procedure,  we train our Wide ResNet-28-2 model on ImageNet downsampled to $32\times32$ \cite{chrabaszcz2017downsampled} (the native image size of CIFAR-10). The top-1 classification error rate is reported for clear demonstration. 

\subsection{Results}
\subsubsection{Results on CUB-200-2011}

\begin{table}[h]
\centering
\begin{tabular}{c|cccc}
\hline
Methods \textbackslash \#label & 2000  & 1000           & 400   & 200   \\ \hline
Supervised labeled            & 68.29 & 53.26          & 28.82 & 17.90 \\ \hline
Pseudo label                  & 71.38 & 49.50          & 25.65 & 10.42 \\
Mean teacher                  & 70.19 & 51.78          & 27.01 & 13.79 \\
MixMatch                      & 73.84 & 60.56          & 32.79 & 22.66 \\
FixMatch                      & 72.76 & 58.30          & 31.03 & 21.86 \\ \hline
AKC                           & 71.33 & 58.42          & 38.71 & 28.57 \\ 
ARC                           & 72.95 & 61.01          & 41.13 & 28.47 \\
\textbf{AKC+ARC}              & 73.65 & 62.01          & 41.69 & 28.96 \\ \hline
\textbf{MixMatch+AKC+ARC}    & 77.51 & 67.26          & 43.80 & 29.55 \\ 
FixMatch+AKC+ARC             & 75.59 & 63.36          & 40.83 & 28.25 \\ \hline
\end{tabular}
\vspace{2pt}
\caption{Classification accuracy of proposed AKC, ARC, and baselines on CUB-200-2011 dataset.}
\label{tab:cub200_result}
\end{table}

The results of adaptive knowledge consistency (AKC),  adaptive representation consistency (ARC), and baseline methods on CUB-200-2011 dataset are listed in \Cref{tab:cub200_result}. The method of combining AKC with ARC achieved best or comparable performance among previous-best baseline methods, especially in the case that labeled samples are fewer. For example, when the size of the labeled dataset is 200, the AKC+ARC method relatively improves the accuracy by 27.8\% compared to MixMatch. One of the advantages of our proposed method is its compatibility. AKC and ARC regularization terms could be combined with other semi-supervised learning methods, like MixMatch and FixMatch. By utilizing AKC and ARC regularization techniques in MixMatch, the performance increased notably. For the fine-tuning with 2000 (and 200) labeled sample, the performance of MixMatch is increased by 5.0\% (and 30.40\%) than vanilla MixMatch. We speculate that one major reason for the effectiveness of AKC and ARC is that AKC and ARC could effectively prevent severe over-fitting when the number of labeled examples is small. \footnote{We notice that FixMatch 
is not superior to MixMatch on CUB-200-2011. 
This observation is partially consistent with the empirical investigation that the benefit of SSL algorithms may be marginal when we transfer the source model to a similar target task  ~\cite{zhou2018semi}. 
}

Results on Indoor-67 are presented in supplementary B.2 .

\subsubsection{Results on MURA}

The results of MURA dataset are listed in \Cref{tab:mura_result}. Although MURA is a medical image dataset with a large domain mismatch from ImageNet, the AKC and ARC can also improve the performance. By utilizing AKC and ARC regularization techniques in FixMatch, the method of FixMatch+AKC+ARC achieves the best performance in both cases of 1000 and 400 labeled samples. 

\begin{table}[h]
\centering
\begin{tabular}{c|cc}
\hline
Methods \textbackslash \#label & 1000  & 400   \\ \hline
Supervised labeled            & 71.95 & 67.54 \\ \hline
Pseudo label                  & 73.99 & 67.56 \\
Mean teacher                  & 72.20 & 65.53 \\
MixMatch                      & 73.85 & 68.94 \\
FixMatch                      & 75.10 & 69.43 \\ \hline
AKC                           & 73.78 & 70.44 \\
ARC                           & 73.91 & 71.19 \\
\textbf{AKC+ARC}              & 73.94 & 71.34 \\ \hline
MixMatch +AKC+ARC             & 74.72 & 70.94 \\
\textbf{FixMatch +AKC+ARC}    & 76.60 & 72.14 \\ \hline
\end{tabular}
\vspace{2pt}
\caption{Classification accuracy of proposed AKC, ARC, and baselines on MURA dataset.}
\label{tab:mura_result}
\vspace{-2mm}
\end{table}

\subsubsection{Results on CIFAR-10}
\begin{table}[h]
\centering
\begin{tabular}{c|ccc}
\hline
Method \textbackslash \#label & 4000 & 250   & 40    \\ \hline
Supervised labeled            & 7.85 & 15.92 & 27.75 \\ \hline
Pseudo label                  & 7.04 & 12.92 & 25.62 \\
Mean teacher                  & 6.43 & 14.03 & 24.67 \\
MixMatch                      & 5.52 & 10.01 & 21.50 \\
FixMatch                      & 4.24 & 5.04  & 9.05  \\ \hline
AKC                           & 6.72 & 14.49 & 24.51 \\
ARC                           & 7.07 & 15.19 & 25.13 \\
\textbf{AKC+ARC}              & 6.55 & 13.93 & 24.17 \\ \hline
MixMatch +AKC+ARC             & 4.92 & 8.95  & 18.90      \\
\textbf{FixMatch +AKC+ARC}    & 4.19 & 4.99  & 7.62      \\ \hline
\end{tabular}
\vspace{2pt}
\caption{Comparison of error rate using proposed AKC, ARC, and baselines on CIFAR-10 dataset.}
\label{tab:CIFAR-10_result}
\end{table}

The results of adaptive knowledge consistency (AKC),  adaptive representation consistency (ARC), and baseline methods on CIFAR-10 dataset are listed in \Cref{tab:CIFAR-10_result}. By utilizing AKC and ARC regularization techniques in FixMatch, the method of FixMatch+AKC+ARC achieves the best performance in both cases of 4000, 250, and 40 labeled samples. By utilizing AKC and ARC regularization techniques in MixMatch, the performance increases notably. When fine-tuning with 250 labeled samples, the error rate of MixMatch is decreased by 10.59\% if we impose AKC and ARC in it. For the previous-best method FixMatch, 
the proposed method still improves the performance, especially in very few labeled data training. 

Note that when 4000 examples (only 8\% of labeled data) are labeled, FixMatch achieves even lower top-1 error rate (4.24\%) than fully supervised learning from scratch using all 50000 examples (5.01\%), indicating that FixMatch employs advanced techniques beyond the mere utilization of unlabeled data. Therefore, it's reasonable that additional improvements will not be remarkable on top of such a competitive baseline. We presented the effectiveness of transfer learning in low-data semi-supervised learning on CIFAR-10 in supplementary B.5. 

\subsubsection{Ablation Study}
Adaptiveness of our proposed AKC and ARC regularization methods is affected by threshold value $\epsilon_{\rm K}$ and $\epsilon_{\rm R}$. If $\epsilon_{\rm K}=0$ (and $\epsilon_{\rm R}=0$), the AKC (and ARC) equals to being removed since non of the sample was selected to calculate the regularization term. If $\epsilon_{\rm K} = {\rm max(H(p_s))} = {\rm log}(C_s) $ (and $\epsilon_{\rm R} = {\rm max(H(p_t))} = {\rm log}(C_t)$), the AKC (and ARC) degenerates to non-adaptive regularization terms with calculating consistency on all samples. 

\begin{table}[h]
\begin{tabular}{c|ccccc}
\hline
$\epsilon_{\rm K}/{\rm log}(C_s) $          & 0     & 0.3   & 0.5   & 0.7   & 1.0   \\ \hline
2000 labels & 68.29 & 70.33 & 70.74 & 71.33 & 70.70 \\
400 labels  &  28.82 & 31.27 & 33.51 & 38.71  & 34.62  \\ \hline
\end{tabular}
\vspace{2pt}
\caption{Performance of proposed AKC under different $\epsilon_{\rm K}$ on CUB-200-2011 dataset with 2000 and 400 labeled samples.}
\label{tab:epsilon_k}
\end{table}

We investigate the performance of AKC under different $\epsilon_{\rm K}$ on CUB-200-2011 dataset, as shown in \Cref{tab:epsilon_k}. As can be seen, AKC achieves better performance with $\epsilon_{\rm K}=0.7 \cdot {\rm log}(C_s)$. This shows the effectiveness of "adaptive" method 
, especially for the case of 400 labeled sample,  adaptive knowledge consistency (with $\epsilon_{\rm K}=0.7 \cdot {\rm log}(C_s)$ ) outperformed standard "non-adaptive" knowledge consistency (with $\epsilon_{\rm K}=  {\rm log}(C_s)$ ) by 11.8\%.

We also investigate the performance of ARC under different $\epsilon_{\rm R}$ on CUB-200-2011 dataset and get similar observations as AKC, as shown in \Cref{tab:epsilon_r}. Thanks to adaptiveness, adaptive representation consistency performs better than non-adaptive representation consistency which uses all samples. In the case of 400 labeled sample, ARC with $\epsilon_{\rm R}=0.5 \cdot {\rm log}(C_t)$ outperforms non-adaptive representation consistency by 5.7\%.

\begin{table}[h]
\begin{tabular}{c|ccccc}
\hline
$\epsilon_{\rm R}/{\rm log}(C_t) $          & 0     & 0.3   & 0.5   & 0.7   & 1.0   \\ \hline
2000 labels & 68.29 & 69.57 & 71.73 & 72.95 & 71.77 \\
400 labels  &  28.82 & 34.01  & 41.88 & 41.13 & 39.63  \\ \hline
\end{tabular}
\vspace{2pt}
\caption{Performance of proposed ARC under different $\epsilon_{\rm R}$ on CUB-200-2011 dataset with 2000 and 400 labeled samples.}
\label{tab:epsilon_r}
\end{table}

\begin{figure}[t]
\begin{center}
\includegraphics[height=0.25\textwidth,width=0.4\textwidth]{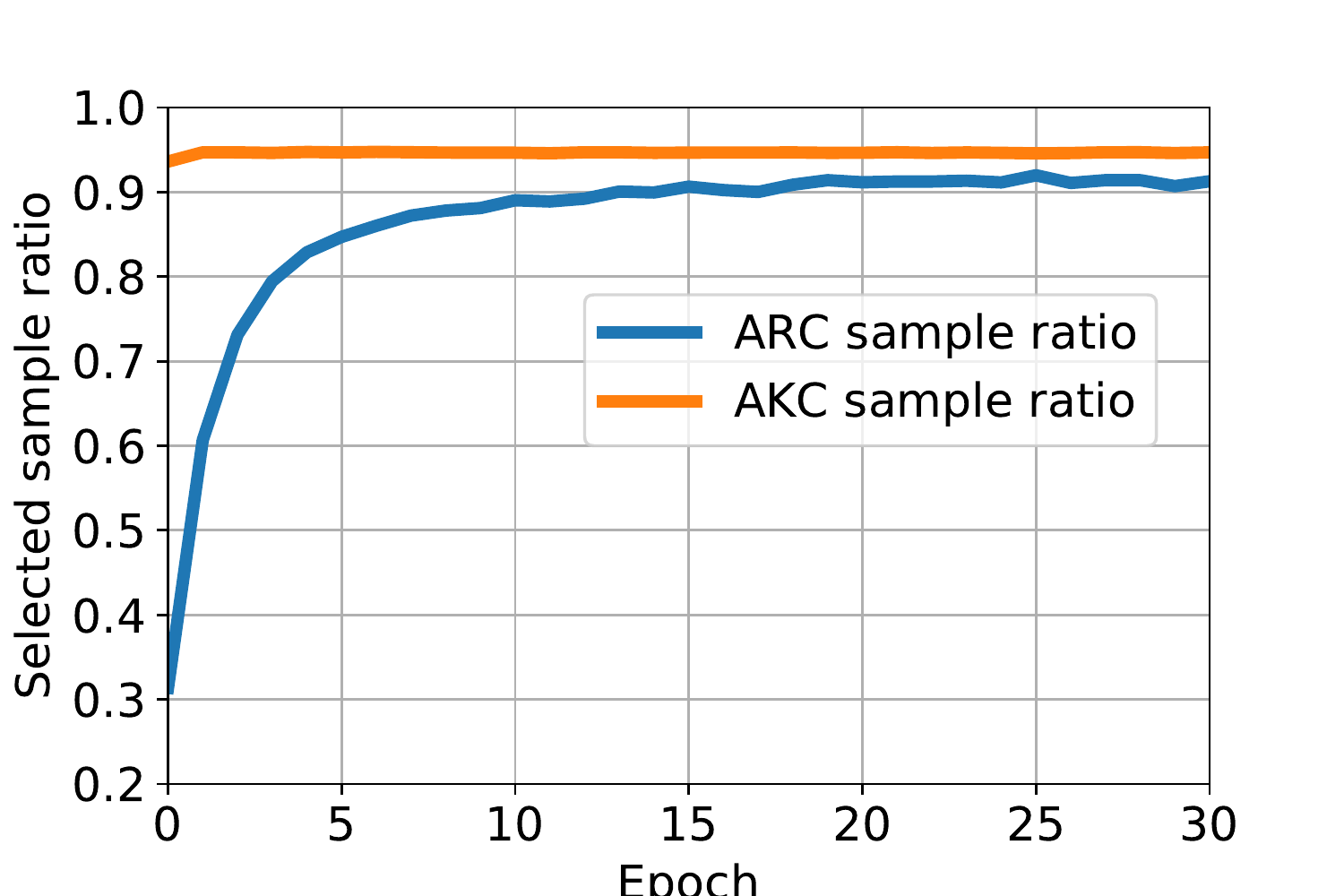}
\caption{Effective sample ratio used in calculating ARC and AKC.}
\label{fig:sample_rate}
\end{center}
\vspace{-2mm}
\end{figure}

The actual sample selected ratio in ARC and AKC is shown in \Cref{fig:sample_rate} on CUB-200-2011 dataset experiment with 2000 labeled samples. As can be seen, the sample selected ratio for ARC is gradually increasing in the first 10 epochs from 0.3 to 0.9. Which can be regarded as a kind of curriculum learning\cite{bengio2009curriculum}. In the earlier stage of training, only a few high confident samples were used for labeled and unlabeled distribution consistency regularization. After 10 epochs of training, the sample ratio converges to near 0.9, indicating that some of the low-confident samples are never used for ARC regularization. This process would be beneficial for training since some "very hard" or abnormal samples might be harmful for generalization. 
The sample selected rate of AKC is stable during training as the source model is frozen during fine-tuning.

Additional experiments on increased accuracy after utilizing AKC and ARC are presented in supplementary B.3.


\subsection{Beyond semi-supervised transfer learning}
Albeit the proposed Adaptive Knowledge Consistency (AKC) and Adaptive Representation Consistency (ARC) regularization methods are targeted at the semi-supervised transfer learning scenario, the application of those two regularization methods is not merely limited to semi-supervised transfer learning tasks.

The AKC regularization can be incorporated with other supervised or unsupervised transfer learning methods since it does not require any label of the target data. Additionally, it is also suitable for tasks which involves multiple models, such as knowledge distillation from
a big teacher model to a small student model.

The ARC regularization is also applicable for semi-supervised learning tasks training from scratch. Additionally, it can also be used in fully supervised learning, where we can easily regard the labeled set as the unlabeled set. \Cref{tab:supervised} shows the result of the AKC and ARC regularization methods in fully supervised transfer learning in CUB-200-2011 dataset. Both AKC and ARC improve the performance of standard transfer learning. 

\begin{table}[]
\centering
\begin{tabular}{c|cccc}
\hline
Method   & Standard & AKC & ARC & ARC+AKC \\ \hline
Accuracy & 81.77    & 82.79 &   82.54   & 83.52 \\ \hline
\end{tabular}
\vspace{2pt}
\caption{Results of AKC and ARC on CUB-200-2011 supervised transfer learning.}
\label{tab:supervised}
\end{table}


%% file: sections/7_appendix.tex
\clearpage
\appendix
\begin{appendices}

\section{Additional Information of Proposed Method}
\label{sec:add_method}

In this paper, we propose two regularization methods: Adaptive Knowledge Consistency (AKC) between the source and target model and Adaptive Representation Consistency (ARC) between labeled and unlabeled examples.

\subsection{Adaptive Knowledge Consistency}
\begin{figure*}[htbp]
\vspace{-3mm}
\begin{center}
\includegraphics[height=0.43\textwidth,width=0.9\textwidth]{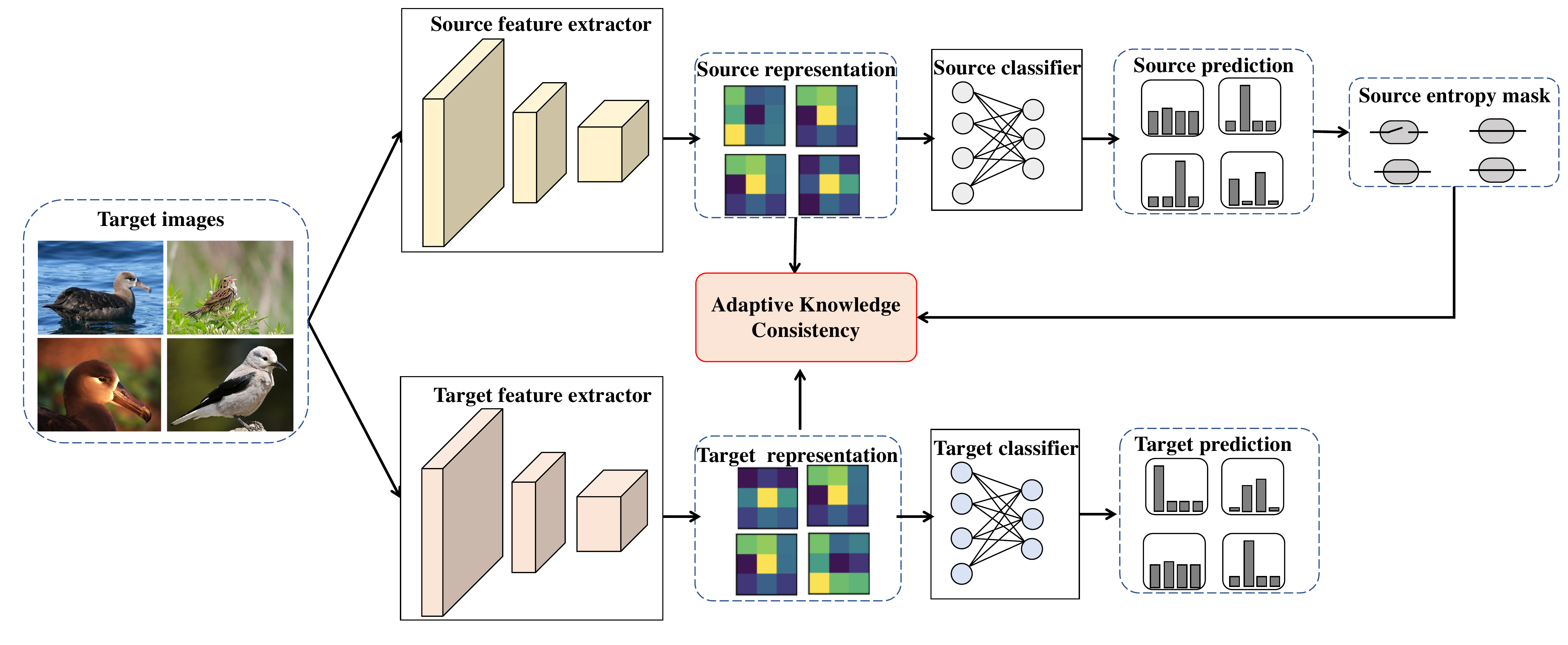}
\caption{Adaptive knowledge consistency between the source and target model.}
\label{fig:kc}
\end{center}
\vspace{-3mm}
\end{figure*}

The AKC regularization can be incorporated with supervised or unsupervised transfer learning methods. As shown in \Cref{fig:kc}, we constrain the  weighted sample-level consistency (Kullback–Leibler divergence or mean square error) of feature-representation between the pre-trained source feature extractor and the target feature extractor using both the labeled and unlabeled samples. The weight of each sample was determined by the entropy of the pre-trained source model's prediction.  

\subsection{Adaptive Representation Consistency}
\begin{figure*}[htbp]
\vspace{-3mm}
\begin{center}
\includegraphics[height=0.6\textwidth,width=0.9\textwidth]{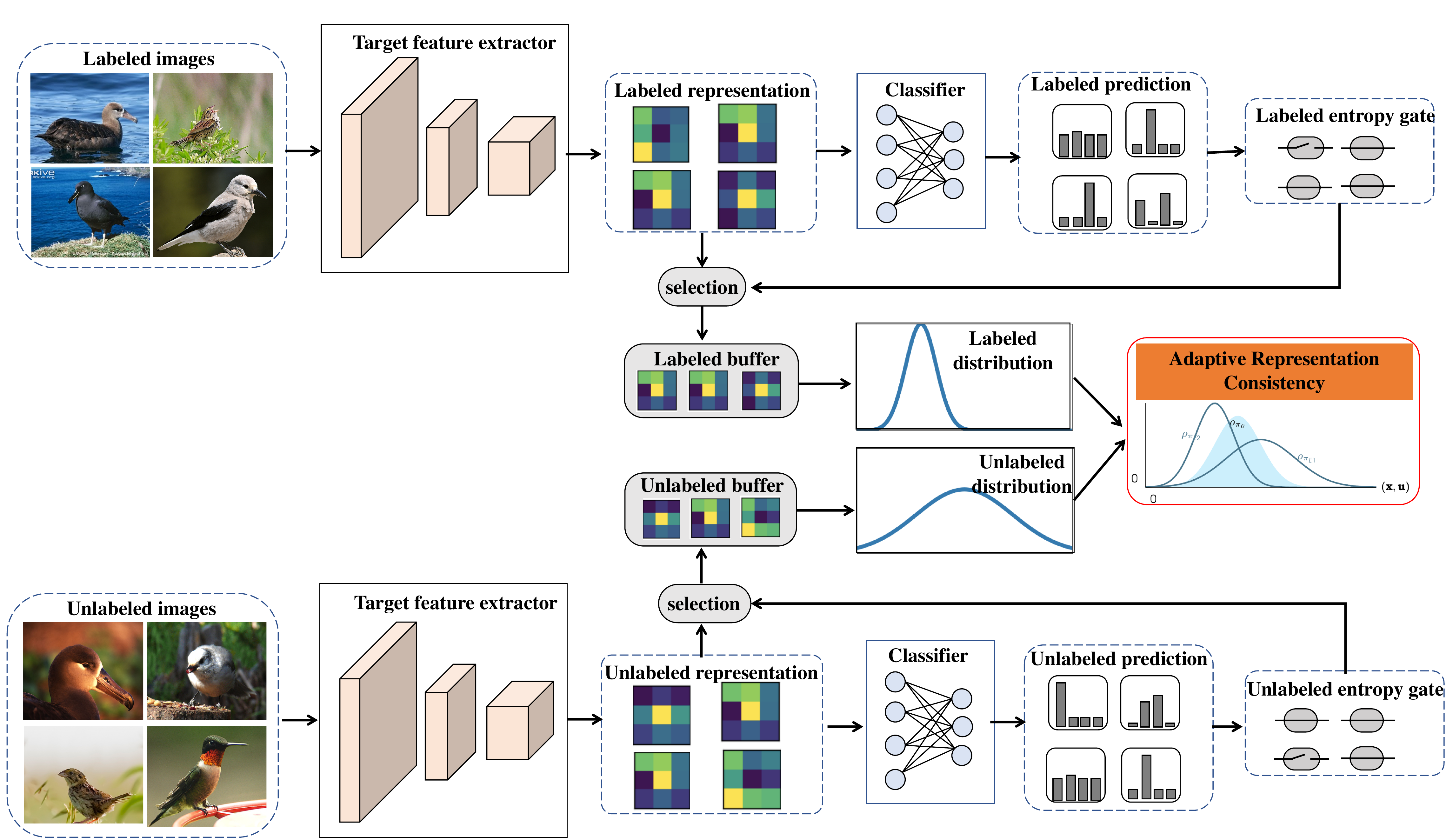}
\caption{Adaptive representation consistency between labeled data distribution and unlabeled data distribution.}
\label{fig:rc}
\end{center}
\vspace{-3mm}
\end{figure*}

The ARC regularization can be used to transfer or learning from scratch semi-supervised methods. As shown in \Cref{fig:rc}, we constrain  Maximum Mean Discrepancies between representations' distribution of selected labeled and selected unlabeled samples. Only confident (labeled and unlabeled) samples with high confidence scores will be selected to regularize the distribution of (labeled and unlabeled) data representation. A high confident sample means that the input sample is more likely to fall into the target model's trust region with low entropy of the prediction. 
To maintain a sufficient number of samples used in ARC regularization, we impose a replay buffer to save recent selected confident samples.

\subsection{Intuitive Explanation of ARC}
\label{sec:intu_arc}

\begin{figure}[h]
\begin{center}
\includegraphics[width=0.4\textwidth]{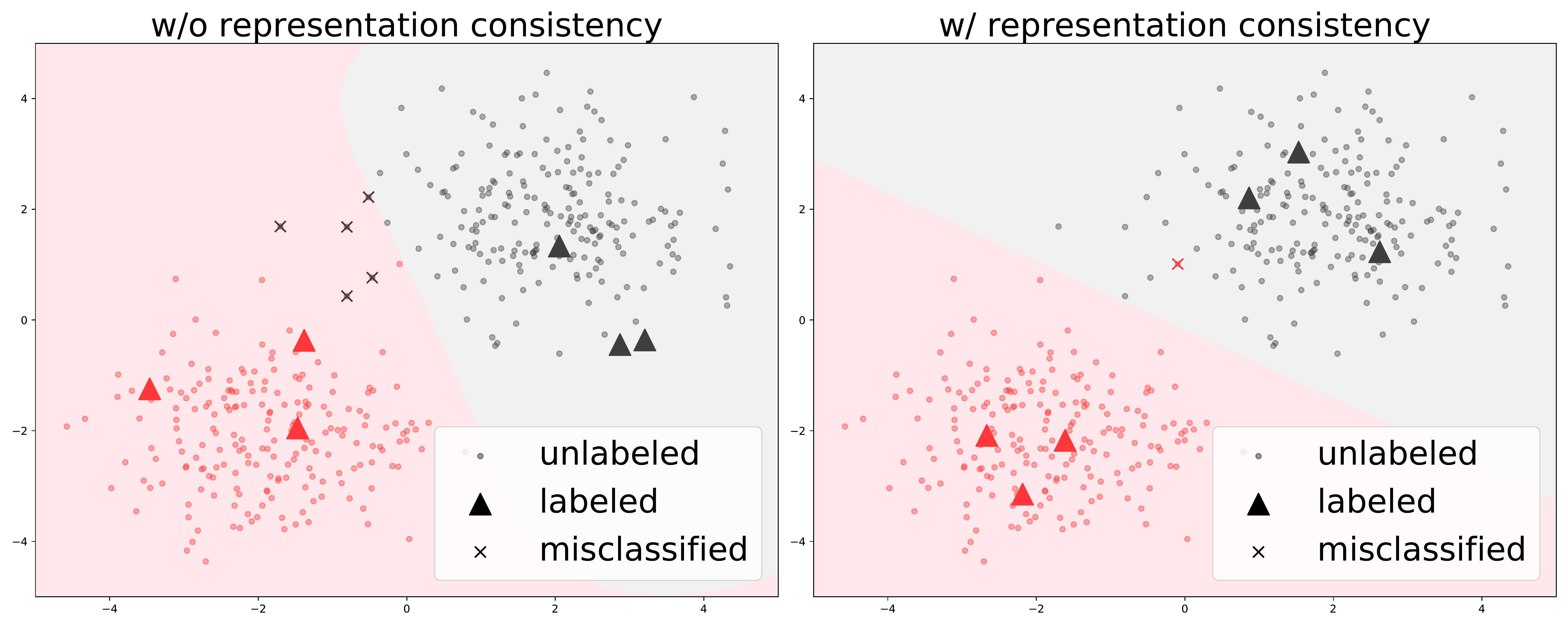}
\caption{Illustration of why enforcing representation consistency helps the model generalize when labeled samples are scarce. Red and black spots denote unlabeled samples. }
\label{fig:intuition_ARC}
\end{center}
\end{figure}

As shown in Figure \ref{fig:intuition_ARC}, although there’s no systematic bias between labeled and unlabeled samples, the risk of sampling bias can be severe when labeled samples are scarce. Without ARC, features learned by unlabeled and labeled data may deviate from each other, but still simultaneously satisfy their constrains due to DNN’s great memorizing capacity. As observed in the plots, this hurts discrimination as misclassification increase even among seen unlabeled samples (left plot), while learned representations induce better decision boundary if labeled samples match the population (right plot).

\begin{table*}[h]
\centering
\begin{tabular}{c|cc|cc|cc}
\hline
\#label      & \multicolumn{2}{c|}{4000} & \multicolumn{2}{c|}{250} & \multicolumn{2}{c}{40}  \\ \hline
Method      & From Scratch  & Transfer  & From Scratch  & Transfer & From Scratch & Transfer \\ \hline
Pseudo label & 16.09         & 7.04      & 49.78         & 12.92    & 79.51            & 25.62    \\
Mean teacher & 9.19          & 6.43      & 32.32         & 14.03    & 74.43            & 24.67    \\
MixMatch     & 6.42          & 5.52      & 11.05         & 10.01    & 47.54        & 21.50    \\
FixMatch     & 4.26          & 4.24      & 5.07          & 5.04     & 13.81        & 9.05     \\ \hline
\end{tabular}
\caption{Comparison of error rate using SSL methods with and without transfer learning.}
\label{tab:ssl_transfer}
\end{table*}

\section{Additional Experiments}
\subsection{Descriptions about Datasets}
\label{sec:dataset}
\begin{itemize}
\setlength{\itemsep}{0pt}

\item CUB-200-2011: The CUB-200-2011 dataset contains 200
fine-grained classes of birds with 11,788 images in total (about 30 images per class for training set and 30 images per class for validation set). In our experiment, we construct the labeled training set with the sample size of $ n \in \{2000, 1000, 400, 200\}$, and use the rest images as unlabeled training set. 

\item MIT Indoor-67: Indoor-67 has 67 scene categories. In each category, there are 80 images for training and 20 images for testing. In our experiment, we construct the labeled training set with the sample size of $ n \in \{1340, 670, 134\}$, and use the rest images as unlabeled training set. 

\item MURA: MURA is a dataset of musculoskeletal radiographs, which contains 40,561 images from 14,863 patient studies. X-ray images are collected from seven parts of human body: elbow, finger, forearm, hand, humerus, shoulder, and wrist. The goal of this dataset is to distinguish normal musculoskeletal studies from abnormal ones (a study often contains more than one image). This
paper follows the experiment setting of \cite{zhou2018semi}: to simply classify normal and abnormal radiographs (one image). For the MURA dataset, We construct the labeled training set with the sample size of $ n \in \{1000, 400\}$, and use the rest images as unlabeled training set.

\item CIFAR-10: The CIFAR-10 dataset is composed of 60,000 images of 10 classes with the size of 32x32. 50,000 images are used for training and 10,000 are used for testing.

\end{itemize}

\begin{table}[h]
\centering
\begin{tabular}{c|ccc}
\hline
Methods \textbackslash \#label & 1340  & 670   & 134   \\ \hline
Supervised labeled            & 68.94 & 63.35 & 44.28 \\ \hline
Pseudo label                  & 71.68 & 63.77 & 39.28 \\
Mean teacher                  & 71.34 & 64.37 & 43.05 \\
MixMatch                      & 73.14 & 68.58 & 44.65  \\
FixMatch                      & 74.27 & 68.31 & 44.13  \\ \hline
AKC                           & 71.93 & 66.64 & 46.79 \\
ARC                           & 72.72 & 66.94 & 46.67 \\
\textbf{AKC+ARC}              & 73.31 & 67.44 & 47.11 \\ \hline
MixMatch +AKC+ARC             & 75.54 & 70.30 & 48.54  \\ 
\textbf{FixMatch +AKC+ARC}   & 76.64  & 70.61 & 48.34  \\ \hline
\end{tabular}
\vspace{2pt}
\caption{Classification accuracy of proposed AKC, ARC, and baselines on Indoor-67 dataset.}
\label{tab:indoor_result}
\end{table}

\subsection{Results on Indoor-67}
\label{sec:indoor}

The experimental results on Indoor-67 dataset are listed in \Cref{tab:indoor_result}. Similar to the results of CUB-200-2011 dataset,  the method of combining AKC with ARC achieves the best or comparable performance among previous-best baseline methods. In the case of 1340 (and 134) labeled sample size, by utilizing AKC and ARC regularization techniques in FixMatch, the performance is increased by 3.2\% (and 9.54\%) than vanilla FixMatch.

\subsection{Empirical study about balancing AKC and ARC}
\label{sec:balance_ac}
\begin{figure}[h]
\begin{center}
\includegraphics[width=0.45\textwidth]{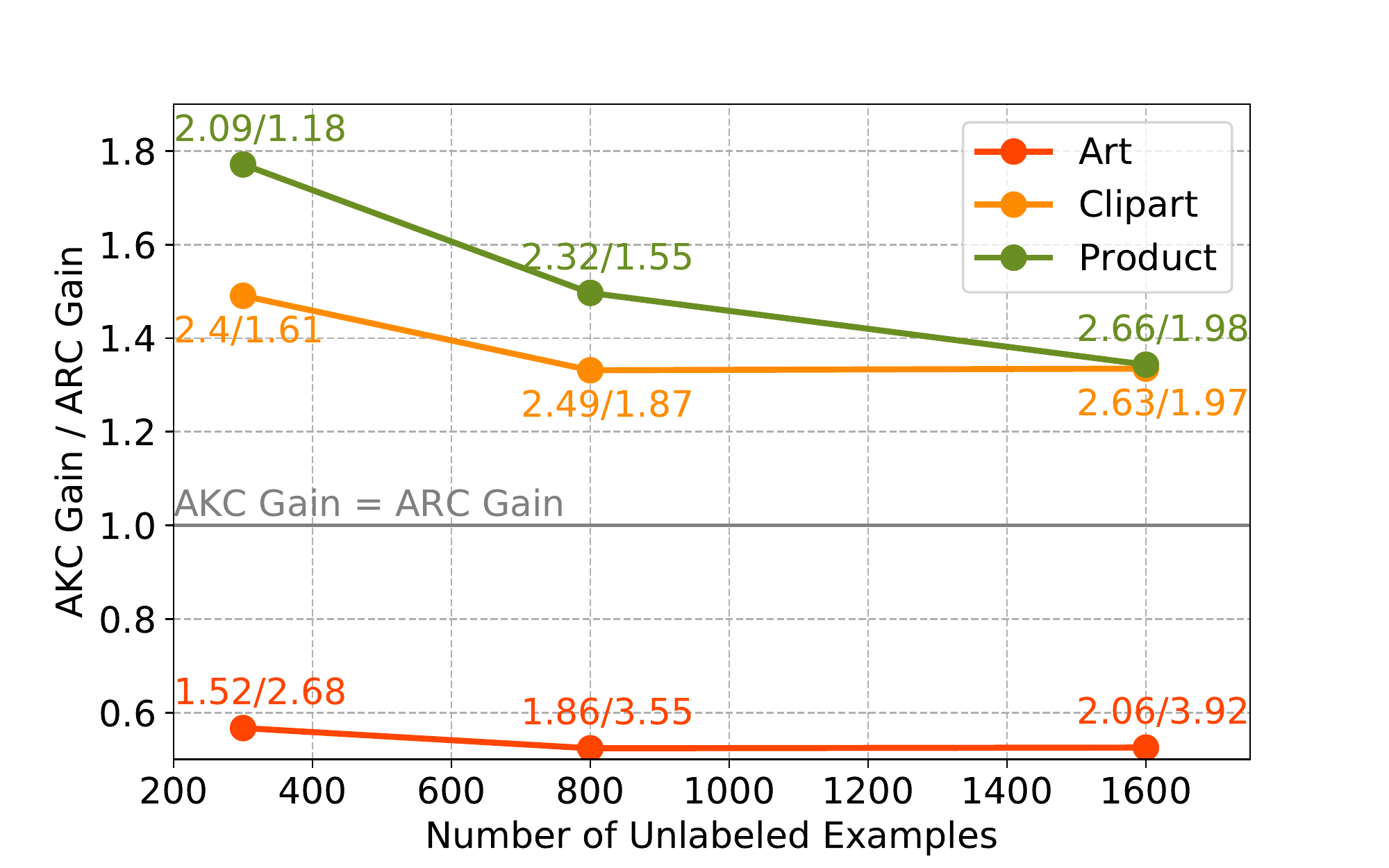}
\caption{Comparison of AKC and ARC gain on Office-Home.}
\label{fig:AKC/ARC}
\end{center}
\end{figure}

We measure the increased accuracy after introducing AKC or ARC on three different Office-Home datasets\footnote{https://www.hemanthdv.org/officeHomeDataset.html}. Generally, as observed in Fig~\ref{fig:AKC/ARC}, AKC is relatively more useful as the discrepancy between the source and target dataset reduces\footnote{\emph{Art} is the most dissimilar with ImageNet due to its particular textures.}, while ARC contributes more with more unlabeled samples provided.

\subsection{The time efficiency of our method}
\label{sec:time_cost}
The proposed AKC and ARC involves almost only extra computation for knowledge distillation in the standard semi-supervised learning framework, which is much more computational efficient than complex operations used in modern SSL methods like MixMatch and FixMatch. Moreover, adding AKC+ARC on top of these competitive methods requires little additional cost as most operations can be reused. For example, combining AKC+ARC and FixMatch only increase 3\% running time compared with the original FixMatch. The actual running time per iteration (in seconds) is measured on CUB-200, as shown in Table \ref{tab:time}.

\begin{table}[!htp]
\small
\begin{tabular}{c|cccccc}
\hline
Method & MM & FM & AKC & ARC & AC & FMAC \\
\hline
Time(s) & 0.629 & 0.563 & 0.531 & 0.513 & 0.562 & 0.580 \\
\hline
\end{tabular}
\caption{Running time per iteration for the CUB-200 experiment evaluated with Tesla V100 GPU. MM: MixMatch, FM: FixMatch, AC: AKC+ARC, FMAC: FM+AC.}
\label{tab:time}
\end{table}

\subsection{Effectiveness of transfer learning in semi-supervised setting \label{sec:ssl_cifar}}
We studied the effectiveness of  transfer learning in some SSL methods on CIFAR-10 dataset, as shown in \cref{tab:ssl_transfer}. As can be seen, transfer learning could considerably improve the performance of SSL methods compared to learning from scratch, especially when labeled examples are insufficient. For example, given only 40 labels, transfer learning improves the performance of the leading SSL method FixMatch by 34.5\% on CIFAR-10.  Thus, the effectiveness of transfer learning in semi-supervised settings was underestimated in the previous works. With the Imprinting technique and proper training strategy, transfer learning could lead to a noticeable improvement.


\end{appendices}

%% file: cvpr.bbl
\begin{thebibliography}{10}\itemsep=-1pt

\bibitem{bengio2009curriculum}
Yoshua Bengio, J{\'e}r{\^o}me Louradour, Ronan Collobert, and Jason Weston.
\newblock Curriculum learning.
\newblock In {\em Proceedings of the 26th annual international conference on
  machine learning}, pages 41--48, 2009.

\bibitem{berthelot2019mixmatch}
David Berthelot, Nicholas Carlini, Ian Goodfellow, Nicolas Papernot, Avital
  Oliver, and Colin~A Raffel.
\newblock Mixmatch: A holistic approach to semi-supervised learning.
\newblock In {\em Proceedings of the Advances in Neural Information Processing
  Systems}, pages 5049--5059, 2019.

\bibitem{borgwardt2006integrating}
Karsten~M Borgwardt, Arthur Gretton, Malte~J Rasch, Hans-Peter Kriegel,
  Bernhard Sch{\"o}lkopf, and Alex~J Smola.
\newblock Integrating structured biological data by kernel maximum mean
  discrepancy.
\newblock {\em Bioinformatics}, 22(14):e49--e57, 2006.

\bibitem{chrabaszcz2017downsampled}
Patryk Chrabaszcz, Ilya Loshchilov, and Frank Hutter.
\newblock A downsampled variant of imagenet as an alternative to the cifar
  datasets.
\newblock {\em arXiv preprint arXiv:1707.08819}, 2017.

\bibitem{cui2018large}
Yin Cui, Yang Song, Chen Sun, Andrew Howard, and Serge Belongie.
\newblock Large scale fine-grained categorization and domain-specific transfer
  learning.
\newblock In {\em Proceedings of the IEEE Conference on Computer Vision and
  Pattern Recognition}, pages 4109--4118, 2018.

\bibitem{deng2009imagenet}
Jia Deng, Wei Dong, Richard Socher, Li-Jia Li, Kai Li, and Li Fei-Fei.
\newblock Imagenet: A large-scale hierarchical image database.
\newblock In {\em Proceedings of the IEEE conference on computer vision and
  pattern recognition}, pages 248--255. Ieee, 2009.

\bibitem{finn2017model}
Chelsea Finn, Pieter Abbeel, and Sergey Levine.
\newblock Model-agnostic meta-learning for fast adaptation of deep networks.
\newblock In {\em International Conference on Machine Learning}, pages
  1126--1135. PMLR, 2017.

\bibitem{fu2019self}
Yang Fu, Yunchao Wei, Guanshuo Wang, Yuqian Zhou, Honghui Shi, and Thomas~S
  Huang.
\newblock Self-similarity grouping: A simple unsupervised cross domain
  adaptation approach for person re-identification.
\newblock In {\em Proceedings of the IEEE/CVF International Conference on
  Computer Vision}, pages 6112--6121, 2019.

\bibitem{furlanello2018born}
Tommaso Furlanello, Zachary Lipton, Michael Tschannen, Laurent Itti, and Anima
  Anandkumar.
\newblock Born again neural networks.
\newblock In {\em International Conference on Machine Learning}, pages
  1607--1616. PMLR, 2018.

\bibitem{ge2017cvpr}
Weifeng Ge and Yizhou Yu.
\newblock Borrowing treasures from the wealthy: Deep transfer learning through
  selective joint fine-tuning.
\newblock In {\em Proceedings of the IEEE conference on computer vision and
  pattern recognition}, pages 10--19, 2017.

\bibitem{gong2013connecting}
Boqing Gong, Kristen Grauman, and Fei Sha.
\newblock Connecting the dots with landmarks: Discriminatively learning
  domain-invariant features for unsupervised domain adaptation.
\newblock In {\em Proceedings of the International Conference on Machine
  Learning}, pages 222--230, 2013.

\bibitem{grandvalet2005semi}
Yves Grandvalet and Yoshua Bengio.
\newblock Semi-supervised learning by entropy minimization.
\newblock In {\em Proceedings of the Advances in neural information processing
  systems}, pages 529--536, 2005.

\bibitem{gretton2012kernel}
Arthur Gretton, Karsten~M Borgwardt, Malte~J Rasch, Bernhard Sch{\"o}lkopf, and
  Alexander Smola.
\newblock A kernel two-sample test.
\newblock {\em The Journal of Machine Learning Research}, 13(1):723--773, 2012.

\bibitem{guo2019spottune}
Yunhui Guo, Honghui Shi, Abhishek Kumar, Kristen Grauman, Tajana Rosing, and
  Rogerio Feris.
\newblock Spottune: transfer learning through adaptive fine-tuning.
\newblock In {\em Proceedings of the IEEE Conference on Computer Vision and
  Pattern Recognition}, pages 4805--4814, 2019.

\bibitem{he2016deep}
Kaiming He, Xiangyu Zhang, Shaoqing Ren, and Jian Sun.
\newblock Deep residual learning for image recognition.
\newblock In {\em Proceedings of the IEEE conference on computer vision and
  pattern recognition}, pages 770--778, 2016.

\bibitem{HintonDistilling}
Geoffrey Hinton, Oriol Vinyals, and Jeff Dean.
\newblock Distilling the knowledge in a neural network.
\newblock {\em stat}, 1050:9, 2015.

\bibitem{huang2007correcting}
Jiayuan Huang, Arthur Gretton, Karsten Borgwardt, Bernhard Sch{\"o}lkopf, and
  Alex~J Smola.
\newblock Correcting sample selection bias by unlabeled data.
\newblock In {\em Proceedings of the Advances in neural information processing
  systems}, pages 601--608, 2007.

\bibitem{huh2016makes}
Minyoung Huh, Pulkit Agrawal, and Alexei~A Efros.
\newblock What makes imagenet good for transfer learning?
\newblock {\em arXiv preprint arXiv:1608.08614}, 2016.

\bibitem{jakubovitz2019lautum}
Daniel Jakubovitz, Miguel~RD Rodrigues, and Raja Giryes.
\newblock Lautum regularization for semi-supervised transfer learning.
\newblock In {\em Proceedings of the IEEE International Conference on Computer
  Vision Workshops}, pages 0--0, 2019.

\bibitem{krizhevsky2009learning}
Alex Krizhevsky, Geoffrey Hinton, et~al.
\newblock Learning multiple layers of features from tiny images.
\newblock 2009.

\bibitem{krizhevsky2012imagenet}
Alex Krizhevsky, Ilya Sutskever, and Geoffrey~E Hinton.
\newblock Imagenet classification with deep convolutional neural networks.
\newblock In {\em Proceedings of the Advances in Neural Information Processing
  Systems}, 2012.

\bibitem{laine2016temporal}
Samuli Laine and Timo Aila.
\newblock Temporal ensembling for semi-supervised learning.
\newblock {\em arXiv preprint arXiv:1610.02242}, 2016.

\bibitem{lee2013pseudo}
Dong-Hyun Lee.
\newblock Pseudo-label: The simple and efficient semi-supervised learning
  method for deep neural networks.
\newblock In {\em Proceedings of the International conference on machine
  learning, Workshop on challenges in representation learning}, volume~3, 2013.

\bibitem{li2018explicit}
Xuhong Li, Yves Grandvalet, and Franck Davoine.
\newblock Explicit inductive bias for transfer learning with convolutional
  networks.
\newblock In {\em 35th International Conference on Machine Learning}, 2018.

\bibitem{li2020rifle}
Xingjian Li, Haoyi Xiong, Haozhe An, Cheng-Zhong Xu, and Dejing Dou.
\newblock Rifle: Backpropagation in depth for deep transfer learning through
  re-initializing the fully-connected layer.
\newblock In {\em International Conference on Machine Learning}, pages
  6010--6019. PMLR, 2020.

\bibitem{li2019delta}
Xingjian Li, Haoyi Xiong, Hanchao Wang, Yuxuan Rao, Liping Liu, and Jun Huan.
\newblock Delta: Deep learning transfer using feature map with attention for
  convolutional networks.
\newblock In {\em International Conference on Learning Representations}, 2018.

\bibitem{li2017learning}
Zhizhong Li and Derek Hoiem.
\newblock Learning without forgetting.
\newblock {\em IEEE transactions on pattern analysis and machine intelligence},
  40(12):2935--2947, 2017.

\bibitem{long2015learning}
Mingsheng Long, Yue Cao, Jianmin Wang, and Michael Jordan.
\newblock Learning transferable features with deep adaptation networks.
\newblock In {\em Proceedings of the International conference on machine
  learning}, pages 97--105. PMLR, 2015.

\bibitem{long2017deep}
Mingsheng Long, Han Zhu, Jianmin Wang, and Michael~I Jordan.
\newblock Deep transfer learning with joint adaptation networks.
\newblock In {\em Proceedings of the 34th International Conference on Machine
  Learning-Volume 70}, pages 2208--2217. JMLR. org, 2017.

\bibitem{loshchilov2016sgdr}
Ilya Loshchilov and Frank Hutter.
\newblock Sgdr: Stochastic gradient descent with warm restarts.
\newblock {\em Learning}, 10:3, 2016.

\bibitem{ngiam2018domain}
Jiquan Ngiam, Daiyi Peng, Vijay Vasudevan, Simon Kornblith, Quoc~V Le, and
  Ruoming Pang.
\newblock Domain adaptive transfer learning with specialist models.
\newblock {\em arXiv preprint arXiv:1811.07056}, 2018.

\bibitem{oliver2018realistic}
Avital Oliver, Augustus Odena, Colin~A Raffel, Ekin~Dogus Cubuk, and Ian
  Goodfellow.
\newblock Realistic evaluation of deep semi-supervised learning algorithms.
\newblock In {\em Proceedings of the Advances in neural information processing
  systems}, pages 3235--3246, 2018.

\bibitem{pan2010domain}
Sinno~Jialin Pan, Ivor~W Tsang, James~T Kwok, and Qiang Yang.
\newblock Domain adaptation via transfer component analysis.
\newblock {\em IEEE Transactions on Neural Networks}, 22(2):199--210, 2010.

\bibitem{pan2009survey}
Sinno~Jialin Pan and Qiang Yang.
\newblock A survey on transfer learning.
\newblock {\em IEEE Transactions on knowledge and data engineering},
  22(10):1345--1359, 2009.

\bibitem{qi2018low}
Hang Qi, Matthew Brown, and David~G Lowe.
\newblock Low-shot learning with imprinted weights.
\newblock In {\em Proceedings of the IEEE conference on computer vision and
  pattern recognition}, pages 5822--5830, 2018.

\bibitem{quattoni2009recognizing}
Ariadna Quattoni and Antonio Torralba.
\newblock Recognizing indoor scenes.
\newblock In {\em Proceedings of the IEEE Conference on Computer Vision and
  Pattern Recognition}, pages 413--420. IEEE, 2009.

\bibitem{rajpurkar2017mura}
Pranav Rajpurkar, Jeremy Irvin, Aarti Bagul, Daisy Ding, Tony Duan, Hershel
  Mehta, Brandon Yang, Kaylie Zhu, Dillon Laird, Robyn~L Ball, et~al.
\newblock Mura dataset: Towards radiologist-level abnormality detection in
  musculoskeletal radiographs.
\newblock {\em Hand}, 1(602):2--215.

\bibitem{saenko2010adapting}
Kate Saenko, Brian Kulis, Mario Fritz, and Trevor Darrell.
\newblock Adapting visual category models to new domains.
\newblock In {\em Proceedings of the European conference on computer vision},
  pages 213--226. Springer, 2010.

\bibitem{shi2009extending}
Yuan Shi, Zhenzhong Lan, Wei Liu, and Wei Bi.
\newblock Extending semi-supervised learning methods for inductive transfer
  learning.
\newblock In {\em Proceedings of the Ninth IEEE international conference on
  data mining}, pages 483--492. IEEE, 2009.

\bibitem{snell2017prototypical}
Jake Snell, Kevin Swersky, and Richard Zemel.
\newblock Prototypical networks for few-shot learning.
\newblock In {\em Proceedings of the Advances in neural information processing
  systems}, pages 4077--4087, 2017.

\bibitem{sohn2020fixmatch}
Kihyuk Sohn, David Berthelot, Nicholas Carlini, Zizhao Zhang, Han Zhang,
  Colin~A Raffel, Ekin~Dogus Cubuk, Alexey Kurakin, and Chun-Liang Li.
\newblock Fixmatch: Simplifying semi-supervised learning with consistency and
  confidence.
\newblock In {\em Proceedings of the Advances in Neural Information Processing
  Systems}, volume~33, 2020.

\bibitem{tarvainen2017mean}
Antti Tarvainen and Harri Valpola.
\newblock Mean teachers are better role models: Weight-averaged consistency
  targets improve semi-supervised deep learning results.
\newblock In {\em Proceedings of the Advances in neural information processing
  systems}, pages 1195--1204, 2017.

\bibitem{torrey2010transfer}
Lisa Torrey and Jude Shavlik.
\newblock Transfer learning.
\newblock In {\em Handbook of research on machine learning applications and
  trends: algorithms, methods, and techniques}, pages 242--264. IGI global,
  2010.

\bibitem{verma2019interpolation}
Vikas Verma, Alex Lamb, Juho Kannala, Yoshua Bengio, and David Lopez-Paz.
\newblock Interpolation consistency training for semi-supervised learning.
\newblock {\em stat}, 1050:19, 2019.

\bibitem{wah2011caltech}
Catherine Wah, Steve Branson, Peter Welinder, Pietro Perona, and Serge
  Belongie.
\newblock The caltech-ucsd birds-200-2011 dataset.
\newblock 2011.

\bibitem{wan2019towards}
Ruosi Wan, Haoyi Xiong, Xingjian Li, Zhanxing Zhu, and Jun Huan.
\newblock Towards making deep transfer learning never hurt.
\newblock In {\em 2019 IEEE International Conference on Data Mining (ICDM)},
  pages 578--587. IEEE, 2019.

\bibitem{wang2020alleviating}
Zhonghao Wang, Yunchao Wei, Rogerio Feris, Jinjun Xiong, Wen-Mei Hwu, Thomas~S
  Huang, and Honghui Shi.
\newblock Alleviating semantic-level shift: A semi-supervised domain adaptation
  method for semantic segmentation.
\newblock In {\em Proceedings of the IEEE/CVF Conference on Computer Vision and
  Pattern Recognition Workshops}, pages 936--937, 2020.

\bibitem{wei2019semi}
Wei Wei, Deyu Meng, Qian Zhao, Zongben Xu, and Ying Wu.
\newblock Semi-supervised transfer learning for image rain removal.
\newblock In {\em Proceedings of the IEEE Conference on Computer Vision and
  Pattern Recognition}, pages 3877--3886, 2019.

\bibitem{yosinski2014transferable}
Jason Yosinski, Jeff Clune, Yoshua Bengio, and Hod Lipson.
\newblock How transferable are features in deep neural networks?
\newblock In {\em Proceedings of the Advances in neural information processing
  systems}, pages 3320--3328, 2014.

\bibitem{yu2020transmatch}
Zhongjie Yu, Lin Chen, Zhongwei Cheng, and Jiebo Luo.
\newblock Transmatch: A transfer-learning scheme for semi-supervised few-shot
  learning.
\newblock In {\em Proceedings of the IEEE/CVF Conference on Computer Vision and
  Pattern Recognition}, pages 12856--12864, 2020.

\bibitem{zagoruyko2016paying}
Sergey Zagoruyko and Nikos Komodakis.
\newblock Paying more attention to attention: Improving the performance of
  convolutional neural networks via attention transfer.
\newblock {\em arXiv preprint arXiv:1612.03928}, 2016.

\bibitem{zagoruyko2016wide}
Sergey Zagoruyko and Nikos Komodakis.
\newblock Wide residual networks.
\newblock In {\em British Machine Vision Conference 2016}. British Machine
  Vision Association, 2016.

\bibitem{zhang2018parameter}
Yinghua Zhang, Yu Zhang, and Qiang Yang.
\newblock Parameter transfer unit for deep neural networks.
\newblock In {\em Pacific-Asia Conference on Knowledge Discovery and Data
  Mining}, pages 82--95. Springer, 2019.

\bibitem{zhou2017places}
Bolei Zhou, Agata Lapedriza, Aditya Khosla, Aude Oliva, and Antonio Torralba.
\newblock Places: A 10 million image database for scene recognition.
\newblock {\em IEEE Transactions on Pattern Analysis and Machine Intelligence},
  2017.

\bibitem{zhou2018semi}
Hong-Yu Zhou, Avital Oliver, Jianxin Wu, and Yefeng Zheng.
\newblock When semi-supervised learning meets transfer learning: Training
  strategies, models and datasets.
\newblock {\em arXiv preprint arXiv:1812.05313}, 2018.

\end{thebibliography}
